\documentclass{Interspeech2024}

\usepackage{tabularx}
\usepackage{balance}
\usepackage{footnote}

\interspeechcameraready 

\title{Enhancing Multilingual Voice Toxicity Detection with Speech-Text Alignment}

\name{Joseph}{Liu$^*$}
\name{Mahesh Kumar}{Nandwana$^*$}
\name{Janne}{Pylkkönen}
\name{Hannes}{Heikinheimo}
\name{Morgan}{McGuire}

\address{Roblox, United States
}
\email{\{josephliu, mnandwana, jpylkkonen, hheikinheimo, morgan\}@roblox.com}


\makeatletter
\newcommand\footnoteref[1]{\protected@xdef\@thefnmark{\ref{#1}}\@footnotemark}
\makeatother
\usepackage{array}
\newcommand{\PreserveBackslash}[1]{\let\temp=\\#1\let\\=\temp}
\newcolumntype{C}[1]{>{\PreserveBackslash\centering}p{#1}}
\newcolumntype{R}[1]{>{\PreserveBackslash\raggedleft}p{#1}}
\newcolumntype{L}[1]{>{\PreserveBackslash\raggedright}p{#1}}

\begin{document}

\maketitle
\def\thefootnote{*}\footnotetext{Equal contribution}
\begin{abstract}
Toxicity classification for voice heavily relies on the semantic content of speech. We propose a novel framework that utilizes cross-modal learning to integrate the semantic embedding of text into a multilabel speech toxicity classifier during training. This enables us to incorporate textual information during training while still requiring only audio during inference. We evaluate this classifier on large-scale datasets with real-world characteristics to validate the effectiveness of this framework. Through ablation studies, we demonstrate that general-purpose semantic text embeddings are rich and aligned with speech for toxicity classification purposes. Conducting experiments across multiple languages at scale, we show improvements in voice toxicity classification across five languages and different toxicity categories.

\end{abstract}

\section{Introduction}


Voice chat is a ubiquitous fixture in online social platforms like gaming and chat rooms, with an increasing number of users frequently utilizing it in a real-time interactive setting. With the growing scale and scope of these platforms, maintaining civility and safety through moderation becomes challenging. Furthermore, for social platforms operating across various geographies, languages, and cultures, moderating voice chat in multiple languages adds further complexity to this issue.
Current approaches for voice toxicity classification use some form of text-based classifier that run on the transcription outputs of automatic speech recognition (ASR)~\cite{ghosh2021detoxy}.
More scalable approaches involve compact classifiers, which classify toxicity directly on audio~\cite{yousefi2021audio, lin2022toxic, nada2023lightweight, adima_gupta}. These methods utilize speech encoders such as Wav2Vec 2.0~\cite{baevski2020wav2vec, nada2023lightweight, adima_gupta} and WavLM~\cite{chen2022wavlm, roblox_mtl}as the base encoder. These approaches enable scalability as this removes an expensive autoregressive ASR component of a high parameter ASR-text classifier in favor of a much smaller and faster non-autoregressive network that directly predicts the classification outputs.

The audio-based speech toxicity classifiers have shown some promise on small-scale speech toxicity datasets, but are severely limited by the availability of large-scale real-world training data. DeToxy~\cite{ghosh2021detoxy} and IEMOCAP~\cite{busso2008iemocap} are public monolingual datasets that are extremely small and lack real-world characteristics given their construction.  For multilingual datasets, availability is much more limited~\cite{costajussà2024mutox}.
Even when researchers have access to extensive internal training datasets like Yousefi and Emmanouilidou~\cite{yousefi2021audio}, the recreation of these datasets on a comparable scale remains challenging due to the substantial human annotation needed for each individual language. 
This approach lacks scalability when aiming to augment training data for toxicity classification across diverse linguistic contexts.

To enhance the performance of audio-based voice toxicity classifiers, various strategies are employed, including modifying the model architecture to better capture the semantic information that distinguishes toxicity. Examples of such strategies include developing a custom attention architecture~\cite{yousefi2021audio}, implementing multi-task learning~\cite{roblox_mtl}, or utilizing pre-trained speech encoders~\cite{baevski2020wav2vec,Duquenne:2023:sonar_arxiv}. One approach even involves feeding speech and text into the classifier at inference time to improve classification performance~\cite{mandal2024attentive}. These methods suggest that forcing the model to learn the semantic information content of speech helps with toxicity classification.
Semantic information in speech has been shown to be learnable in a fixed embedding space that can span multiple languages and modalities, and can be incorporated into speech processing tasks~\cite{NEURIPS2021_multimodal_multilingual, blau23_interspeech}. Machine Translation architectures such as ConST~\cite{ye2022cross} have already benefited from this phenomenon to improve performance in their domain.

Inspired by advances in cross-modal learning, this paper proposes a cross-modal framework for training a pre-trained speech encoder by leveraging a text-encoder on a multilabel toxicity detection task. Through this framework, we contribute the following: Firstly, we demonstrate that a strong text encoder can improve the performance of a toxicity classifier by injecting text during training. Secondly, we present empirical results on large-scale monolingual and multilingual datasets with real-world characteristics, highlighting the feasibility of this method for use in real-world applications. Thirdly, we show that toxicity classification inherently reduces to a semantic speech representation problem, by applying text injection at different layers of the network, demonstrating that learning a simple linear projection of a robust embedding space is sufficient for the model to show improvement over the baseline. Finally, we show that this framework can improve an audio-only voice toxicity classifier for multiple languages at scale without requiring text input during inference time.


\section{Data Pipeline}
Data labeling for voice toxicity is a slow and expensive process. It also exposes human annotators to toxic content, which ideally should be minimized. We developed a fully automatic scalable data labeling pipeline that labels audio with toxicity classes, and text transcripts which is used for text injection during training, eliminating the need for human annotation.

The data pipeline is the same as described in our previous work~\cite{roblox_mtl}, with a three stage process which includes splitting an audio file into chunks using the Opus codec's DTX extension, followed by using publicly available ASR models~\cite{whispermodelpaper_icml} to transcribe the data, before using a text toxicity classifier which is built on DistilBERT~\cite{sanh2019distilbert} to label the data with toxicity labels\footnote{https://research.roblox.com/publications/how-we-scaled-bert-to-serve-1-billion-daily-requests-on-cpu}. During the ASR stage, we further break down the audio chunks into sentences as determined by the ASR model output. 
We only use ASR during the data labeling phase, and not during training or inference in this paper.
We allocate our limited resources for human annotation for validation and test sets, where each audio sample is labeled by multiple annotators to reduce human error. 

\section{Proposed System}

The proposed system consists of two main components: a speech toxicity classifier, which consists primarily of a speech encoder, and a second component, a multilingual text encoder.
\subsection{Model Architecture}
We use a publicly available speech encoder~\cite{whispermodelpaper_icml} for all our experiments in this paper, with the English-only small\footnote{https://huggingface.co/openai/whisper-small.en} variant being used for the monolingual experiments, and the multilingual small\footnote{https://huggingface.co/openai/whisper-small} variant being used for the multilingual experiments. We use the Multilingual E5 Large encoder\footnote{https://huggingface.co/intfloat/multilingual-e5-large}~\cite{wang2024multilingual} as the text encoder for all our experiments, as it is one of the best performing multilingual text encoders according to the Massive Text Embedding Benchmark (MTEB)~\cite{muennighoff-etal-2023-mteb}.


\begin{figure}[t]
  \centering
  \includegraphics[width=\linewidth]{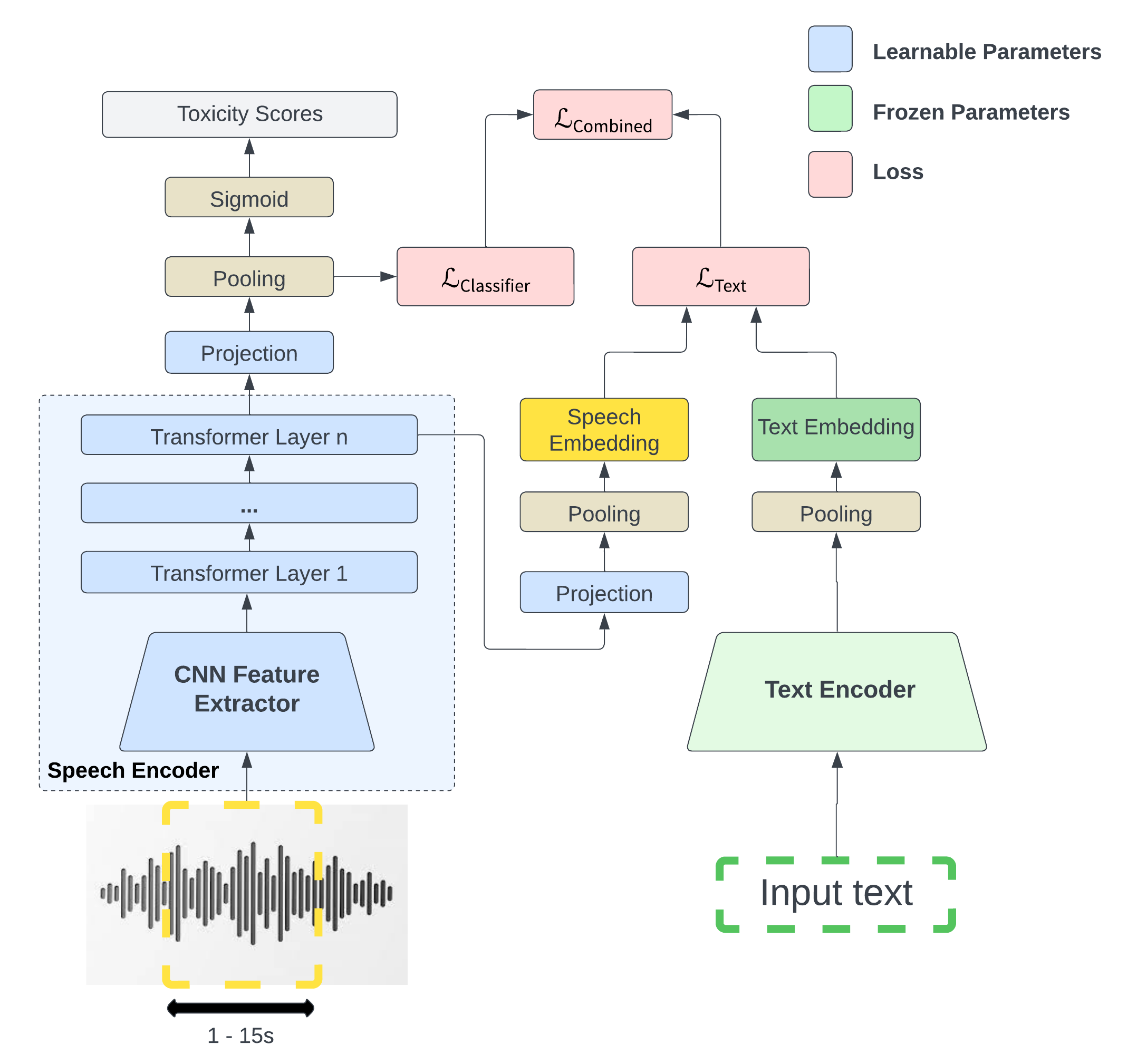}
  \caption{Schematic diagram of proposed text injection training pipeline.}
  \label{fig:proposed_system}
  \vspace{-2mm}
\end{figure}

\subsection{Speech Toxicity Classifier}
The Toxicity Classifier comprises six distinct categories: Profanity, Bullying, Dating \& Sexting, Racism, Other, and No-Violation. The Other category encompasses a diverse range of toxic speech, including references to Grooming, Drugs and Alcohol, Radicalization, and more, which don't neatly fit into the initial four toxic categories detected by the text toxicity classifier. As a single audio clip can contain multiple types of violations, the classification task is a multi-label rather than a multi-class classification problem. The network consists of a speech encoder which feeds into a projection then pooling layer before classification. The classifier is then finetuned with all layers unfrozen with Binary Cross Entropy (BCE) loss.  During inference, we only run the speech toxicity classifier.
\subsection{Text Encoder}
The pretrained text encoder, used in this work, is trained on diverse text content to produce a rich semantic embedding space. Prior literature reports that robust multilingual sentence embeddings~\cite{multilingual_sentence10.1162/tacl_a_00288} can be used in downstream tasks such like speech translation, as demonstrated in frameworks like ConST~\cite{ye2022cross} and SONAR~\cite{Duquenne:2023:sonar_arxiv}. 
Inspired by these, the sentence embedding of the audio transcript can be injected into the speech toxicity classifier training process to augment the speech encoder's semantic understanding of the speech. The text encoder is frozen and only used during training.

\subsection{Text Injection Framework}
The proposed training framework as shown in Figure \ref{fig:proposed_system} is formulated as a linear combination of two losses, primarily the classifier loss $\mathcal{L}_{\text{classifier}}$ and the text injection loss $\mathcal{L}_{\text{text}}$, with hyperparameter $\alpha$. The classifier loss is computed on the predictions of the toxicity classifier, while the text injection loss forces alignment of the output of the text encoder to the speech encoder. The combined loss $\mathcal{L}_{\text{combined}}$ is defined as:
\begin{equation}
    \mathcal{L}_{\text{combined}} = \alpha  \mathcal{L}_{\text{classifier}} + (1-\alpha) \mathcal{L}_{\text{text}}
\end{equation}
For this paper, BCE is used for $\mathcal{L}_{\text{classifier}}$ in all experiments as it is suitable for the multilabel formulation.
Two losses of key interest for prior success in other tasks involving cross-modal alignment are the Mean-Squared Error (MSE) loss and the multi-class N-pair contrastive loss~\cite{NIPS2016_metric_loss}. MSE has shown promise in the topic of knowledge distillation~\cite{kim_mse_kd}, as well as for cross-modal encoders like SONAR~\cite{Duquenne:2023:sonar_arxiv}, while the multi-class N-pair contrastive loss.~\cite{NIPS2016_metric_loss} has shown to work for the cross-modal training of speech and text machine translation~\cite{ye2022cross}.
These losses are then applied at some layer in the speech encoder network. Given a targeted speech encoder layer $i$ as function $\ell_i$, the speech $s$ and the corresponding transcript of the audio $\textit{t}$, we define the two variables for both losses:
\begin{align}
    x &= \text{MeanPool}(h_\text{Proj}(\ell_i(s))) \label{eq:1} \\
    y &= \text{MeanPool}(\textit{t}) \label{eq:2}
\end{align}
where $h_\text{Proj}$ is the learnable projection layer applied to the speech encoder layer outputs to align mismatched dimension sizes between the layer outputs and the text encoder outputs.
The mean squared error for text injection loss we use in this paper is calculated as:
\begin{align}
    \mathcal{L}_{\text{MSE}} = ||x-y||^{2}
\end{align}
We use contrastive loss as defined in Sohn et al.~\cite{NIPS2016_metric_loss} and Ye et al.~\cite{ye2022cross}, where for each speech segment $s$ and their corresponding transcript $y$ in a training batch of $N$ examples, we randomly pick a set of $N-1$ transcripts $\{t_{i}^{-}\}^{N-1}_{i=1}$
\begin{align}
    \mathcal{L}_{\text{Contrastive}} = - \sum_{s,t}{\log{\frac{\exp(sim(x,y)/\tau)}{\sum_{t_j\in \mathcal{A}}{\exp(sim(x,y(t_j))/\tau}}}}
\end{align}
where $\mathcal{A} = {t} \cup \{t_{i}^{-}\}^{N-1}_{i=1}$ (the set of all transcripts in a given training batch), $\tau$ is a temperature hyperparameter, $sim$ is the cosine similarity function $sim(a,b)=\frac{a^\intercal b}{||a||||b||}$. 
Both losses are defined to align the output of the text encoder to the speech encoder, and are only used during training in this framework.

\section{Experiments}
We demonstrate the text injection framework in both monolingual and multilingual settings. We first highlight in the monolingual setting the optimal loss hyperparameter $\alpha$, and investigate ablations for different losses, as well as understanding how injecting text at different layers of the model influences the final performance of the model. We then show that this approach generalizes in a multilingual setting.
\subsection{Training protocol}
We use the following training protocol for all the experiments to ensure equitable comparison and reproducibility. All speech encoders were finetuned from a pretrained checkpoint. We use Adam optimizer with a learning rate (LR) of 3.0$\times 10^{-5}$, an epsilon value of 1$\times 10^{-8}$ and a weight decay factor of 0.1. We use a linear warmup LR scheduler with a warmup rate of 0.1. $\tau$ for contrastive loss was not tuned in this paper, and defaulted to 1. Given that contrastive loss is sensitive to batch sizes as loss is computed across all batch samples, we use a per-device batch size of 8 with gradient accumulation of 4 for an effective per-device batch size of 32, across 8 A100 GPUs, for a cumulative batch size of 256 for all experiments. Each experiment is trained for 6 epochs, which takes approximately 15 hours for the English dataset and 42 hours for the multilingual dataset per run. For all experiments, the final output of the text encoder is used for text injection into the speech encoder.


To study different aspects of the text injection framework, we use monolingual experiments, which we train on English speech data. Multilingual experiments demonstrate the scalability of the optimal text injection framework across multiple languages. By integrating a shared multilingual semantic embedding into the training process, the model achieves generalization across diverse languages. The multilingual model maintains the same architecture as the monolingual model, utilizing a single unified classifier head for toxicity classification across all languages. Preliminary findings indicate that both the speech and text encoders must be pretrained on the same languages as those used for fine-tuning in toxicity classification, as the framework does not directly address language-specific feature extraction. The baseline used for all experiments is specifically the same model trained without introducing text information (with $\alpha = 1$) on the same datasets, similar in concept to the baseline used in our previous work~\cite{roblox_mtl}.

For evaluation, we evaluate audio on 15 second chunks of audio at a time. For audio longer than 15 seconds, we use a stride of 15 seconds to divide it into smaller chunks, which are then aggregated together per audio sample using the maximum class-wise probability of all 15-second frames to compute metrics. We pick the best performing model per experiment based on the validation set, and only report the test set results. The primary evaluation metric we use is per-class average precision (AP). To compute confidence intervals for this metric, we bootstrap it using the ConfidenceIntervals library\footnote{Ferrer, L. and Riera, P. Confidence Intervals for evaluation in machine learning [Computer software]. https://github.com/luferrer/ConfidenceIntervals}, at a 95\% confidence interval with 1000 bootstrap sets.

\begin{table}[t]
\centering
\vspace{-2mm}
\fontsize{6.5}{7.5}\selectfont
\setlength{\tabcolsep}{1.75pt}
\vspace{1mm}
\caption{Training and evaluation data statistics.}
\begin{tabularx}{1.0\columnwidth}{C{10mm}|C{7mm}|C{7.5mm}C{7.5mm}C{7.5mm}C{7.5mm}C{7.5mm}C{7.5mm}C{7.5mm}}
\toprule
  \tiny Language & \tiny Label Type & \tiny Profanity & \tiny Dating~\& Sexting & \tiny Racist & \tiny Bullying & \tiny Other & \tiny No Violation & \tiny Total \\
\midrule
\multicolumn{9}{c}{\bf Training Dataset (hours)} \\
\midrule
 English &\tiny Pipeline & 1,755.1 & 307.2 & 225.0 & 902.5 & 1,559.5 & 1,791.4 & 4,080.3 \\
 Spanish  &\tiny Pipeline& 400.6 & 295.7 & 40.7 & 165.4 & 255.5 & 997.1 & 1,653.4 \\
French  &\tiny Pipeline& 451.2 & 226.0 & 26.5 & 204.2 & 246.8 & 1,014.0 & 1,644.3 \\
 German  &\tiny Pipeline& 155.2 & 75.6 & 15.5 & 86.3 & 98.6 & 371.0 & 615.2 \\
 Portuguese  & \tiny Pipeline& 961.0 & 842.2 & 152.2 & 425.9 & 431.0 & 2,384.3 & 3,840.3 \\
 \midrule
 \multicolumn{9}{c}{\bf  Validation Dataset (hours)} \\
\midrule
 English & \tiny Human & 10.9 & 1.1 & 1.1 & 2.8 & - & 79.3 & 91.9 \\
 Spanish & \tiny Human & 32.0 & 5.1 & 3.5 & 15.9 & -& 129.9 & 166.7 \\ 
 French & \tiny Human & 11.3 & 2.6 & 1.0 & 5.5 & - & 53.6 & 67.7 \\ 
 German  & \tiny Pipeline & 13.7 & 6.1 & 1.8 & 8.3 & 9.9 & 261.4 & 280.7 \\ 
 Portuguese & \tiny Pipeline & 88.0 & 78.9 & 15.6 & 43.1 & 44.0 & 1,148.1 & 1270.4 \\ 
\midrule
\multicolumn{9}{c}{\bf  Human-Labeled Test Dataset (hours)} \\
\midrule
 English & \tiny Human & 316.37 & 57.32 & 53.75 & 73.01 & - & 140.11 & 493.27 \\ 
 Spanish & \tiny Human & 45.67 & 12.86 & 5.06 & 23.95 & - & 98.86 & 152.88 \\ 
 French & \tiny Human & 40.60 & 5.84 & 2.89 & 19.36 & - & 41.87 & 88.13 \\ 
 German & \tiny Human & 7.23 & 1.08 & 1.19 & 3.84 & - & 32.03 & 40.71 \\ 
 Portuguese & \tiny Human & 32.17 & 12.27 & 11.06 & 18.60 & - & 10.85 & 45.93 \\ 
\bottomrule
\end{tabularx}
\label{tab:data_duration_statistics}
\vspace{-5mm}
\end{table}

\begin{figure*}[t!]
  \centering
  \includegraphics[width=\linewidth]{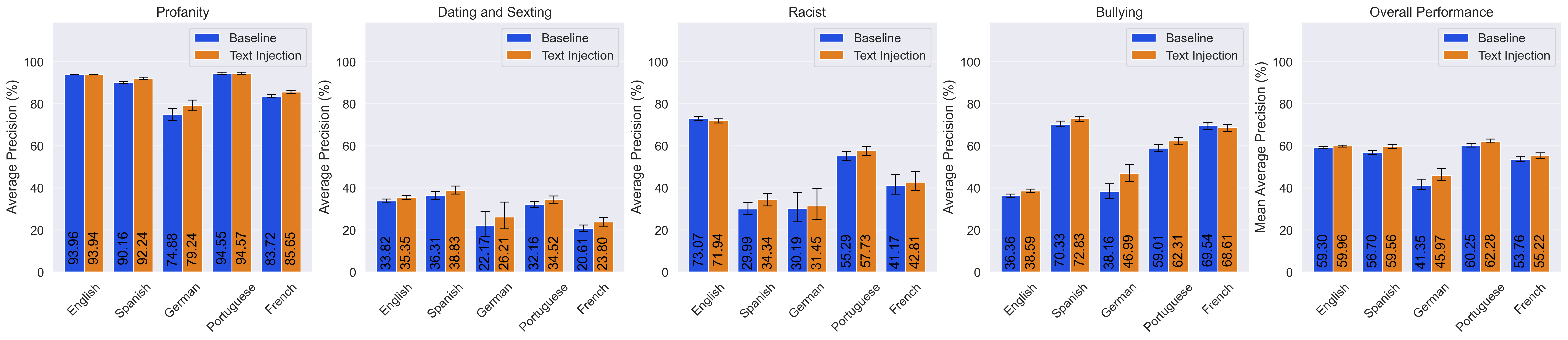}
  \vspace{-7mm}
  \caption{Results comparing multilingual text injection against baseline across 5 languages along with their 95\% confidence intervals.}
  \label{fig:multilingual_results}
\vspace{-0mm}
\end{figure*}

\subsection{Training \& Evaluation Datasets}
For training, we train on data generated and labeled from our data pipeline. Human-labeled datasets are only labeled with the toxicity labels, and are used for the validation and test sets. For German and Portuguese, due to the lack of human labels, we substitute the human label set with a validation set labeled using the data pipeline. Training, validation and test datasets are disjoint. We report the per language breakdown for each multilingual dataset. Table \ref{tab:data_duration_statistics} shows the given breakdown of every dataset used in this paper. All data is sourced from real-world voice chat data.
Audio segments selected for training span from 1 to 15 seconds, which are segmented by sentence. Audio less than 1 second is discarded due to the lack of meaningful semantic context to determine toxicity of that sample. For evaluation, audio segments selected span from 1 to 60 seconds, segmented by Opus's DTX extension.

\section{Results}
In this section, we investigate different aspects of the text injection pipeline, and show the corresponding ablations on the English monolingual toxicity classification task, as well as the relative performance on the multilingual toxicity classification task. 
While the model is trained on 6 categories, we report the average precision (AP) on the four toxicity categories, and the mAP across the 4 different toxicity categories and compute the mAP, as the "Other" category is a catch-all for other less prevalant form of toxicity that is an output of the text classifier during the labeling process but not labeled by humans. We report only the results from the human-labeled test set.

\subsection{Monolingual}
We perform ablations on English only dataset. All of the experiments in Table \ref{tab:monolingual_ablations} had the text injection loss applied on the final output layer of the speech encoder of the toxicity classifier, while experiments in Table \ref{tab:monolingual_ablations_layers} are performed on a speech encoder with 12 transformer layers. Preliminary observations of Table \ref{tab:monolingual_ablations}'s baseline suggest more keyword based categories like Profanity and Racist have much higher average precision across all models versus more semantic based categories like Bullying and Dating \& Sexting. For nearly all experimental groups, there is a statistically significant improvement over the baseline across most categories for both investigated losses.

\begin{table}[t]
\centering
\vspace{-1mm}
\caption{Ablations for different losses and input parameters.}
\fontsize{6.75}{7.75}\selectfont
\setlength{\tabcolsep}{2pt}
\vspace{-1mm}
\begin{tabularx}{1\columnwidth}{C{11mm}|C{10mm}|C{10mm}C{10mm}C{10mm}C{10mm}C{10mm}}
\toprule
   Loss &   $\alpha$ weight &  Profanity & Dating~\& Sexting & Racist & Bullying  & mAP \\
\midrule
Baseline & 1 & 93.80 & 35.39 & 66.26 & 37.15 & 58.15 \\
\midrule
MSE & 0.1 & 94.73 & 37.19 & 73.65 & 40.67 & 61.56 \\
MSE & 0.5 & 93.75 & 35.98 & 68.50 & 38.29 & 59.13 \\
MSE & 0.9 & 94.18 &	35.11 &	69.30 &	38.75 &	59.33 \\ 
 \midrule
Contrastive & 0.1 & 95.35 & 37.90 & 77.25 & 40.84 & 62.84  \\
Contrastive & 0.5 & 95.14 & 37.19 & 73.28 & 37.81 & 60.85 \\
Contrastive & 0.9 & 93.41 & 35.35 & 73.87 & 38.20 & 60.21 \\
\bottomrule
\end{tabularx}
\label{tab:monolingual_ablations}
\vspace{-4mm}
\end{table}
\subsubsection{Text Injection Loss}
Table \ref{tab:monolingual_ablations} shows that contrastive loss performs better across multiple toxicity categories, with particularly stronger improvements observed in semantic heavy categories like Dating \& Sexting, Racist and Bullying. We note that in particular, the Racist category sees an almost 10\% improvement in the best performing contrastive model. This implies that the best performing model is learning a robust internal representation via contrastive loss, an observation consistent with results in machine translation~\cite{ye2022cross}. As using MSE can be seen as a form of knowledge distillation~\cite{kim_mse_kd} from the text to speech domain, there is an implication that the embedding of the text encoder has a domain gap with the speech encoder. This is not observed when using contrastive loss as the speech encoder learns its own internal representation without explicit numerical alignment to the text embedding, being able to learn a representation that is more suitable to the domain of toxicity classification. In both losses though, we see how aligned the text and audio modalities are for the purposes of toxicity classification, and this suggests cross-modal methods can be a path to handle low-resource scenarios.

\subsubsection{Optimal Loss Hyperparameters}
Ablations in Table \ref{tab:monolingual_ablations} show that lower values of $\alpha$ yields better performing toxicity classifiers, suggesting that the toxicity classification task depends on a strong semantic representation of audio, and that getting a robust representation of audio and classifying toxic speech are mostly aligned objectives. This suggests for the toxicity classification task that even the baseline speech toxicity classifiers are learning speech embeddings with a similar internal representation to general purpose text embeddings used for natural language processing tasks. 

\subsubsection{Optimal Layer Selection}
Table \ref{tab:monolingual_ablations_layers} shows that for the last output layer is the most ideal layer to apply contrastive learning on, with the greatest gains applied on the last (12th) layer. This corroborates our hypothesis in the previous ablations that the final representation of the speech encoder is most aligned to the text encoder, as no additional nonlinear transformations were needed to select task-specific features from the text encoder for optimal performance. 

\begin{table}[t]
\centering
\vspace{-1mm}
\caption{Ablations for Optimal Layer Selection on a 12 layer model}
\fontsize{6.75}{7.75}\selectfont
\setlength{\tabcolsep}{2pt}
\vspace{-1mm}
\begin{tabularx}{1\columnwidth}{C{8mm}|C{11.5mm}|C{10mm}C{10mm}C{10mm}C{10mm}C{10mm}}
\toprule
    Layer &  Loss & Profanity & Dating~\& Sexting & Racist & Bullying  & mAP \\
\midrule
 1 &  Contrastive & 94.74 & 35.53 & 72.33 & 37.28 & 59.97\\
 6 &  Contrastive & 94.75 & 35.70 & 73.04 & 36.52 & 60.00\\
 12 &  Contrastive& 95.35 & 37.90 & 77.25 & 40.84 & 62.84\\
\bottomrule
\end{tabularx}
\label{tab:monolingual_ablations_layers}
\vspace{-6mm}
\end{table}

\subsection{Multilingual}

We see in Figure \ref{fig:multilingual_results} that even in the baseline scenario, a encoder with a single output head is able to perform multilingual speech toxicity classification. There is a statistically significant notable improvement over the baseline models across almost all languages and categories, with relative improvements up to 23\% for Bullying in German, and 11\% overall. As with the monolingual setting, we notice that the most noticable are in toxicity categories that are less keyword-based and more semantic focused, like Dating and Sexting, Racism and Bullying. We do note that this benefit seems to be more pronounced in lower-resource languages with less training data. Since the encoder was jointly trained with multiple languages, the benefit could be attributed to the multilingual text embedding allowing generalization across multiple languages. This provides evidence that this framework works in a multilingual setting and at scale.


\section{Conclusion}

We present a framework that introduces text into the training pipeline for an audio-based toxicity classifier. Our experiments suggests introducing semantic knowledge from the text domain can improving audio-only toxicity classification. Ablations show the strong alignment between text semantic embeddings and the task of audio toxicity classification, while highlighting the minor domain gap between speech and text embeddings as evidenced by the improved performance of contrastive loss over MSE loss. We also prove that this method can scale to multiple languages, and perform on real-world data orders of magnitude above those reported in prior literature. In the future, we plan to investigate other methods of introducing cross-modal alignment, including jointly training a speech and text toxicity classifier. It is our hope that this work improves our understanding about speech toxicity detection at scale and provides some insights useful to the research community about this task.

\balance
\bibliographystyle{IEEEtran}
\bibliography{mybib}

\end{document}